\documentclass[10pt,twocolumn,letterpaper]{article}

\renewcommand\footnotemark{}
\usepackage{cvpr}
\usepackage{times}
\usepackage{epsfig}
\usepackage{graphicx}
\usepackage{amsmath}
\usepackage{amssymb}
\usepackage{textcomp}
\usepackage{subfigure}
\usepackage{mathtools}

\usepackage{multirow}
\usepackage{enumitem}
\newcommand{\figref}[1]{Fig. \ref{#1}}
\newcommand{\tabref}[1]{Table \ref{#1}}
\newcommand{\equref}[1]{(\ref{#1})}

\usepackage[pagebackref=true,breaklinks=true,letterpaper=true,bookmarks=false]{hyperref}

\cvprfinalcopy 

\def\cvprPaperID{9411} 

\ifcvprfinal\fi
\begin{document}

\title{Cylindrical Convolutional Networks for\\Joint Object Detection and Viewpoint Estimation\thanks{This research was supported by R\&D program for Advanced Integrated-intelligence for Identification (AIID) through the National Research Foundation of KOREA (NRF) funded by Ministry of Science and ICT (NRF-2018M3E3A1057289).}}

\author{Sunghun Joung$^{1}$, Seungryong Kim$^{2,3}$, Hanjae Kim$^{1}$, Minsu Kim$^{1}$,\\ Ig-Jae Kim$^{4}$, Junghyun Cho$^{4}$, and Kwanghoon Sohn$^{1,*}$\thanks{$^{*}$Corresponding author}\\
$^1$Yonsei University
$^2$\'Ecole Polytechnique F\'ed\'erale de Lausanne (EPFL)\\
$^3$Korea University
$^4$Korea Institute of Science and Technology (KIST)\\
{\tt\small{\{sunghunjoung,incohjk,minsukim320,khsohn\}@yonsei.ac.kr}}\\
\tt\small{seungryong\_kim@korea.ac.kr},
{\tt\small{\{drjay,jhcho\}@kist.re.kr}}
}

\maketitle

\begin{abstract}
Existing techniques to encode spatial invariance within deep convolutional neural networks only model 2D transformation fields. This does not account for the fact that objects in a 2D space are a projection of 3D ones, and thus they have limited ability to severe object viewpoint changes. To overcome this limitation, we introduce a learnable module, cylindrical convolutional networks (CCNs), that exploit cylindrical representation of a convolutional kernel defined in the 3D space. CCNs extract a view-specific feature through a view-specific convolutional kernel to predict object category scores at each viewpoint. With the view-specific feature, we simultaneously determine objective category and viewpoints using the proposed sinusoidal soft-argmax module. Our experiments demonstrate the effectiveness of the cylindrical convolutional networks on joint object detection and viewpoint estimation.
\end{abstract}
\vspace{-6pt}

\section{Introduction}\label{sec:1}
Recent significant success on visual recognition, such as image classification \cite{Simonyan2014}, semantic segmentation \cite{Long2015}, object detection \cite{Girshick2014}, and instance segmentation \cite{He2017}, has been achieved by the advent of deep convolutional neural networks (CNNs). Their capability of handling geometric transformations mostly comes from the extensive data augmentation and the large model capacity \cite{Krizhevsky2012,He2016,Sabour2017}, having limited ability to deal with severe geometric variations, e.g., object scale, viewpoints and part deformations.
To realize this, several modules have been proposed to explicitly handle geometric deformations.
Formally, they transform the input data by modeling spatial transformation \cite{Jaderberg2015,Choy2016,Lin2017inverse}, e.g., affine transformation, or by learning the offset of sampling locations in the convolutional operators \cite{Yi2016,Dai2017}.
However, all of these works only use a visible feature to handle geometric deformation in the 2D space, while viewpoint variations occur in the 3D space.
\begin{figure}[!t]
\centering
{\includegraphics[width=0.99\linewidth]{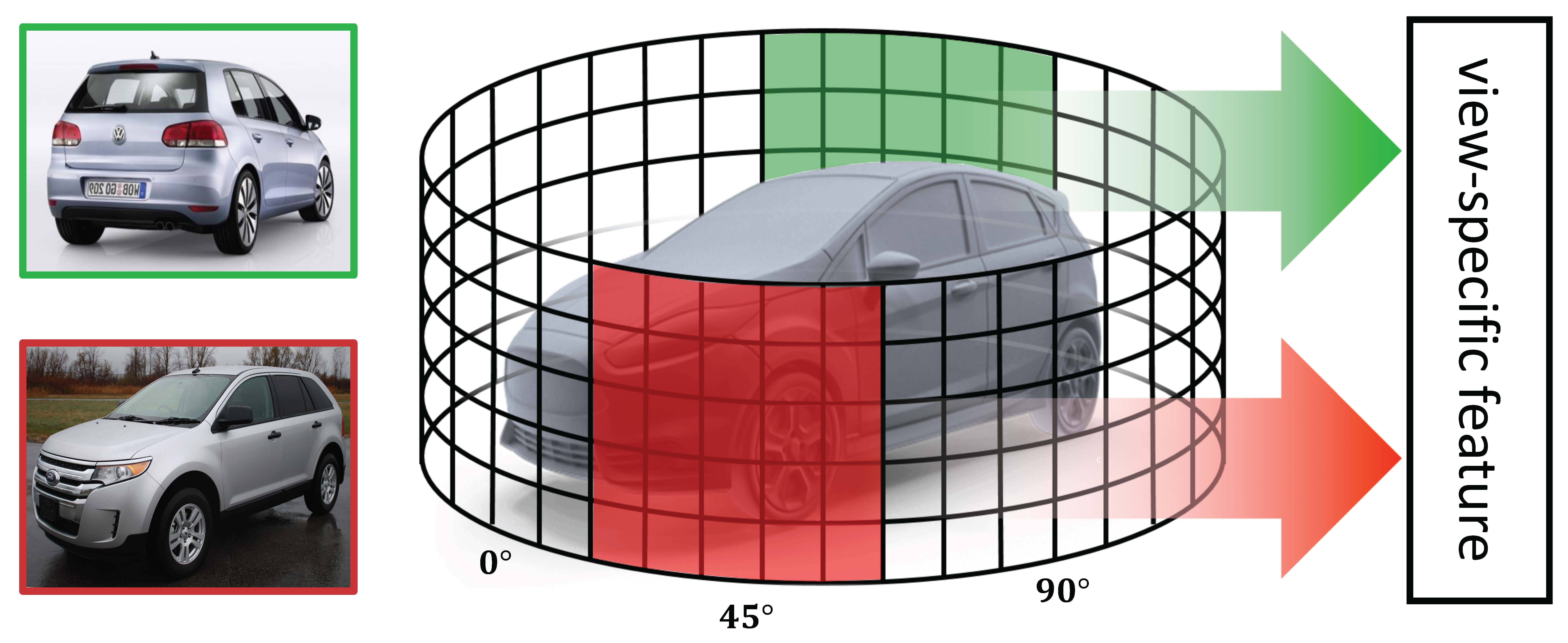}}\\
\vspace{4pt}
\caption{Illustration of cylindrical convolutional networks (CCNs) : Given a single image of objects, we apply a view-specific convolutional kernel to extract the shape characteristic of object from different viewpoints.}
\label{fig:1}\vspace{-10pt}
\end{figure}

To solve the problems of viewpoint variations, joint object detection and viewpoint estimation using CNNs \cite{Tulsiani2015,Su2015,Massa2016,Divon2018} has recently attracted the interest.
This involves first estimating the location and category of objects in an image, and then predicting the relative rigid transformation between the camera coordinate in the 3D space and each image coordinate in the 2D space.
However, category classification and viewpoint estimation problems are inherently contradictory, since the former requires a \emph{view-invariant} feature representation while the latter requires a \emph{view-specific} feature representation.
Therefore, incorporating viewpoint estimation networks to a conventional object detector in a multi-task fashion does not help each other, as demonstrated in several works \cite{Massa2016,Elhoseiny2016}.

Recent studies on 3D object recognition have shown that object viewpoint information can improve the recognition performance. Typically, they first represent a 3D object with a set of 2D rendered images, extract the features of each image from different viewpoints, and then aggregate them for object category classification \cite{Su2015multi,Bai2016,Wang2017dominant}.
By using multiple features with a set of predefined viewpoints, they effectively model shape deformations with respect to the viewpoints.
However, in real-world scenarios, they are not applicable because we cannot access the invisible side of an object without 3D model.

In this paper, we propose cylindrical convolutional networks (CCNs) for extracting view-specific features and using them to estimate object categories and viewpoints simultaneously, unlike conventional methods that share representation of feature for both object category \cite{Ren2015,Liu2016,Lin2017} and viewpoint estimation \cite{Su2015,Massa2016,Divon2018}.
As illustrated in \figref{fig:1}, the key idea is to extract the view-specific feature conditioned on the object viewpoint (i.e., azimuth) that encodes structural information at each viewpoint as in 3D object recognition methods \cite{Su2015multi,Bai2016,Wang2017dominant}.
In addition, we present a new and differentiable argmax operator called sinusoidal soft-argmax that can manage sinusoidal properties of the viewpoint to predict continuous values from the discretized viewpoint bins.
We demonstrate the effectiveness of the proposed cylindrical convolutional networks on joint object detection and viewpoint estimation task, achieving large improvements on Pascal 3D+ \cite{Xiang2014} and KITTI \cite{Geiger2012} datasets.
\section{Related Work}\label{sec:2}
\begin{figure*}[!t]
    \centering
    \subfigure[]{\includegraphics[height=0.195\textheight]{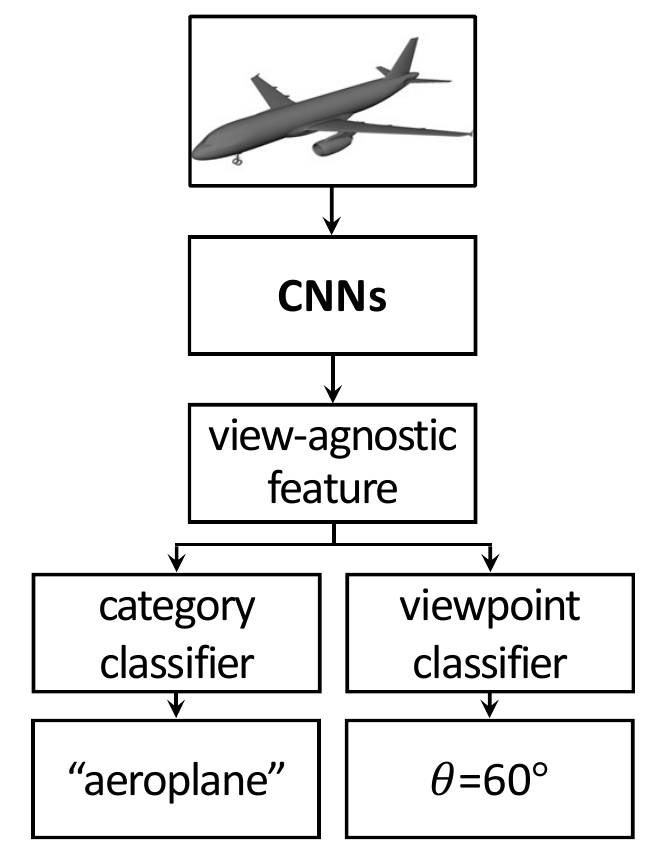}}\hfill
    \subfigure[]{\includegraphics[height=0.195\textheight]{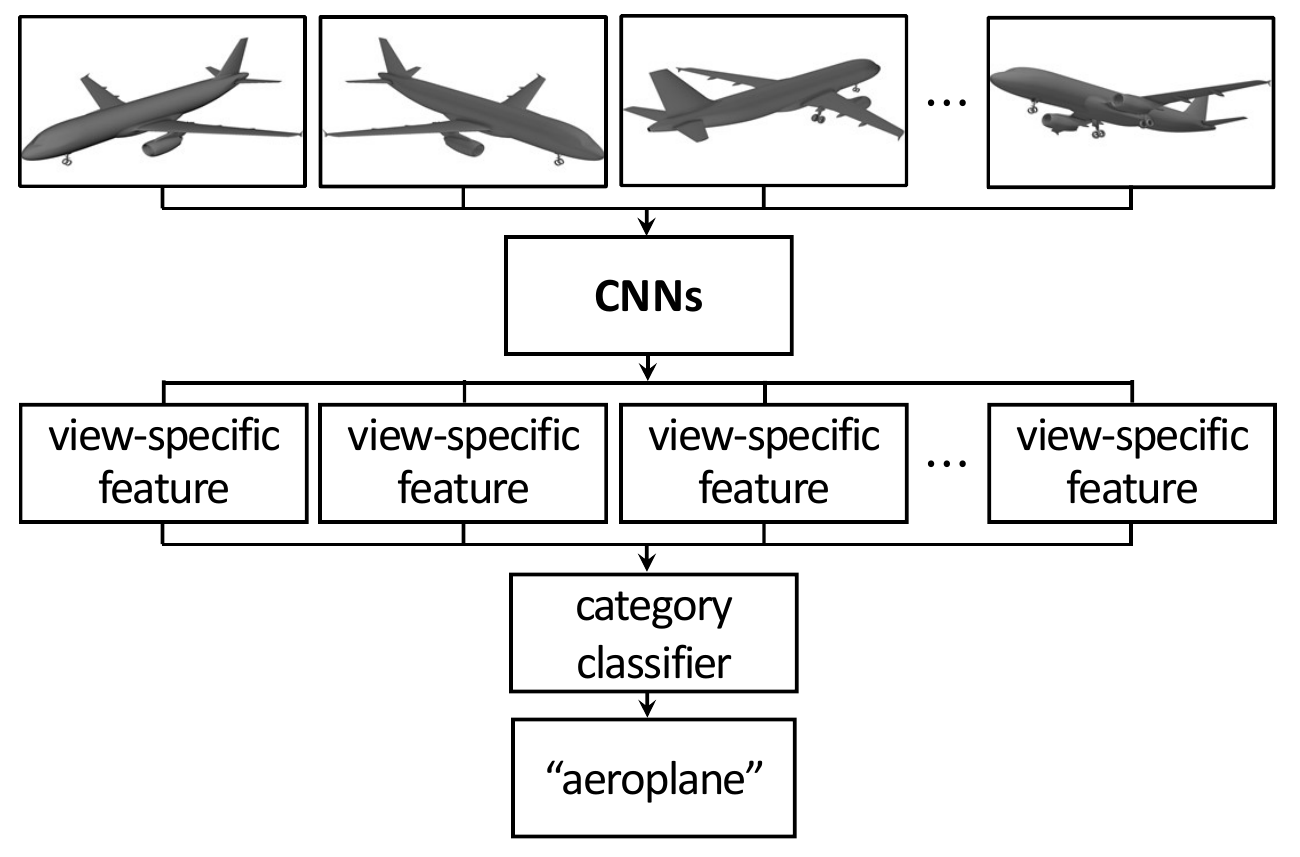}}\hfill
    \subfigure[]{\includegraphics[height=0.195\textheight]{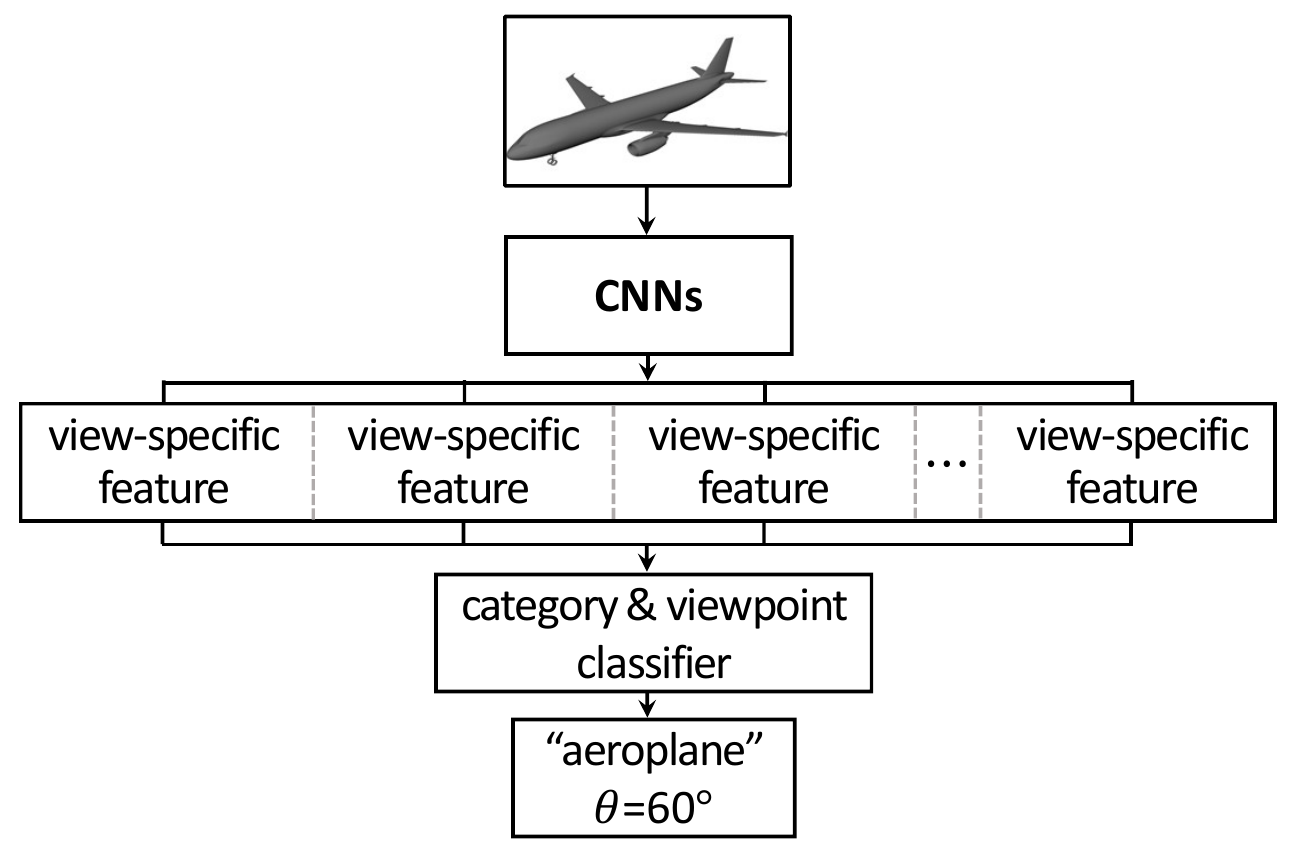}}\hfill
    \vspace{2pt}
    \caption{Intuition of cylindrical convolutional networks: (a) joint category and viewpoint estimation methods \cite{Massa2016,Divon2018} using single-view image as an input, (b) 3D object recognition methods \cite{Su2015multi,Bai2016} using multi-view image as an input, and (c) cylindrical convolutional networks, which take the advantages of 3D object recognition methods by extracting view-specific features from single-view image as an input.}
    \label{fig:2}\vspace{-10pt}
\end{figure*}
\paragraph{2D Geometric Invariance.}\label{sec:21}

Most conventional methods for visual recognition using CNNs \cite{Simonyan2014,Girshick2014,Long2015} provided limited performance due to geometric variations.
To deal with geometric variations within CNNs, spatial transformer networks (STNs) \cite{Jaderberg2015} offered a way to provide geometric invariance by warping features through a global transformation.
Lin and Lucey \cite{Lin2017inverse} proposed inverse compositional STNs that replace the feature warping with transformation parameter propagation, but it has a limited capability of handling local transformations.
Therefore, several methods have been introduced by applying convolutional STNs for each location \cite{Choy2016}, estimating locally-varying geometric fields \cite{Yi2016}, and estimating spatial transformation in a recursive manner \cite{Kim2018}.
Furthermore, to handle adaptive determination of scales or receptive field for visual recognition with fine localization, Dai \etal \cite{Dai2017} introduced two new modules, namely, deformable convolution and deformable ROI pooling that can model geometric transformation for each object.
As all of these techniques model geometric deformation in the projected 2D image only with visible appearance feature, there is a lack of robustness to viewpoint variation, and they still only rely on extensive data augmentation.
\vspace{-10pt}

\paragraph{Joint Category and Viewpoint Estimation.}\label{sec:22}
Since viewpoint of 3D object is a continuous quantity, a natural way to estimate it is to setup a viewpoint regression problem.
Wang \etal \cite{Wang2016} tried to directly regress viewpoint to manage the periodic characteristic with a mean square loss.
However, the regression approach cannot represent the ambiguities well that exist between different viewpoints of objects with symmetries or near symmetries \cite{Massa2016}.
Thus, other works \cite{Tulsiani2015,Su2015} divide the angles into non-overlapping bins and solve the prediction of viewpoint as a classification problem, while relying on object localization using conventional methods (i.e. Fast R-CNN \cite{Girshick2015}).
Divon and Tal \cite{Divon2018} further proposed a unified framework that combines the task of object localization, categorization, and viewpoint estimation.
However, all of these methods focus on accurate viewpoint prediction, which does not play a role in improving object detection performance \cite{Massa2016}.

Another main issue is a scarcity of real images with accurate viewpoint annotation, due to the high cost of manual annotation.
Pascal 3D+ \cite{Xiang2014}, the largest 3D image dataset still is limited in scale compare to object classification datasets (e.g. ImageNet \cite{Deng2009}).
Therefore, several methods \cite{Su2015,Wang2016,Divon2018} tried to solve this problem by rendering 3D CAD models \cite{Chang2015} into background images, but they are unrealistic and do not match real image statistics, which can lead to domain discrepancy.
\vspace{-10pt}
\paragraph{3D Object Recognition.}\label{sec:23}
There have been several attempts to recognize 3D shapes from a collection of their rendered views on 2D images.
Su \etal \cite{Su2015multi} first proposed multi-view CNNs, which project a 3D object into multiple views and extract view-specific features through CNNs to use informative views by max-pooling.
GIFT \cite{Bai2016} also extracted view-specific features, but instead of pooling them, it obtained the similarity between two 3D objects by view-wise matching.
Several methods to improve performance have been proposed, by recurrently clustering the views into multiple sets \cite{Wang2017dominant} or aggregating local features through bilinear pooling \cite{Yu2018}.
Kanezaki \etal \cite{Kanezaki2018} further proposed RotationNet, which takes multi view images as an input and jointly estimates object's category and viewpoint.
It treats the viewpoint labels as latent variables, enabling usage of only a partial set of multi-view images for both training and testing.
\section{Proposed Method}\label{sec:3}
\subsection{Problem Statement and Motivation}\label{sec:31}
Given a single image of objects, our objective is to jointly estimate object category and viewpoint to model viewpoint variation of each object in the 2D space.
Let us denote ${N}_{c}$ as the number of object classes, where the class $C$ is determined from each benchmark and ${N}_{v}$ is determined by the number of discretized viewpoint bins.
In particular, since the variation of elevation and tilt is small on real-scenes \cite{Xiang2014}, we focus on estimation of the azimuth.

Object categorization requires a view-agnostic representation of an input so as to recognize the object category regardless of viewpoint variations.
In contrast, viewpoint estimation requires a representation that preserves shape characteristic of the object in order to distinguish their viewpoint.
Conventional CNNs based methods \cite{Massa2016,Divon2018} extract a \emph{view-agnostic} feature, followed by task-specific sub-networks, i.e., object categorization and viewpoint estimation, as shown in \figref{fig:2} (a). They, however, do not leverage the complementary characteristics of the two tasks, thus showing a limited performance. 
Unlike these methods, some methods on 3D object recognition have shown that \emph{view-specific} features for each viewpoint can encode structural information \cite{Su2015multi,Bai2016}, and thus they use these feature to facilitate the object categorization task as shown in \figref{fig:2} (b).
Since they require multi-view images of pre-defined viewpoints, their applicability is limited to 3D object recognition (i.e. ModelNet 40 \cite{Wu2015}).

To extract the \emph{view-specific} features from a single image, we present cylindrical convolutional networks that exploit a cylindrical convolutionial kernel, where each subset is a view-specific kernel to capture structural information at each viewpoint.
By utilizing view-specific feature followed by object classifiers, we estimate an object category likelihood at each viewpoint and select a viewpoint kernel that predicts to maximize object categorization probability.

\begin{figure*}[!t]
\centering
{\includegraphics[width=0.99\linewidth]{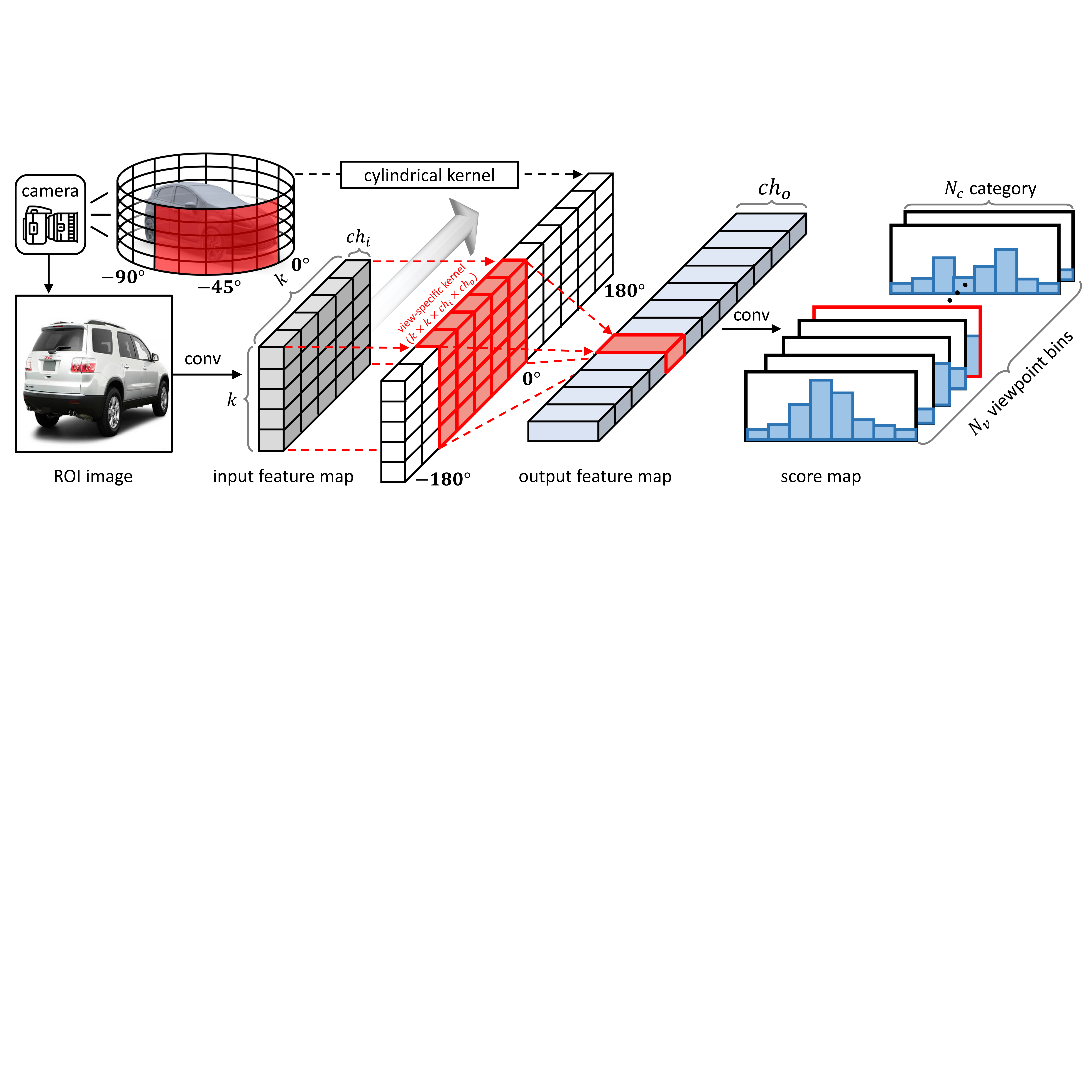}}\\
\vspace{8pt}
\caption{Key idea of cylindrical convolutional networks. Input feature maps from fully convolutional networks are fed into the cylindrical convolutional kernel to obtain ${N}_{v}$ variants of view-specific feature.
Then, each view-specific feature is used to identify its category likelihood that object category classification and viewpoint estimation can be jointly estimated.}
\label{fig:3}\vspace{-10pt}
\end{figure*}

\subsection{Cylindrical Convolutional Networks}\label{sec:32}
Let us denote an intermediate CNN feature map of Region of Interest (ROI) \cite{He2017} as $\mathrm{x}\in{\mathbb{R}}^{k\times k\times {\mathrm{ch}}_{i}}$, with spatial resolution $k\times k$ and ${\mathrm{ch}}_{i}$ channels.
Conventional viewpoint estimation methods \cite{Massa2016,Divon2018} apply a $k\times k$ view-agnostic convolutional kernel in order to preserve position sensitive information for extracting feature $F\in{\mathbb{R}}^{{\mathrm{ch}}_{o}}$, where ${\mathrm{ch}}_{o}$ is the number of output channels.
Since the structural information of projected images varies with different viewpoints, we aim to apply a view-specific convolutional kernel at a predefined set of ${N}_{v}$ viewpoints. 
The most straightforward way for realizing this is to define ${N}_{v}$ variants of $k\times k$ kernel. This strategy, however, cannot consider structural similarity between nearby viewpoints, and would be inefficient.

We instead model a cylindrical convolutional kernel with weight parameters ${\mathop{W}}_{\mathrm{cyl.}}\in{\mathbb{R}}^{k\times {N}_{v}\times {\mathrm{ch}}_{i}\times{\mathrm{ch}_{o}}}$ as illustrated in \figref{fig:3}.
Each $k\times k$ kernel extracted along horizontal axis on ${\mathop{W}}_{\mathrm{cyl.}}$ in a sliding window fashion can be seen as a view-specific kernel ${\mathop{W}}_{\mathrm{v}}$.
We then obtain ${N}_{v}$ variants of a view-specific feature ${F}_{v}\in{\mathbb{R}}^{\mathrm{ch}_{o}}$ as
\begin{equation}\label{equ:convolution}
{F}_{v}=\sum\limits_{\mathrm{p}\in{\mathcal{R}}}{{\mathop{W}}_{\mathrm{v}}\left(\mathrm{p}\right)\cdot\mathrm{x}\left(\mathrm{p}\right)}=\sum\limits_{\mathrm{p}\in{\mathcal{R}}}{{\mathop{W}}_{\mathrm{cyl.}}\left(\mathrm{p}+{\mathrm{o}}_{v}\right)\cdot\mathrm{x}\left(\mathrm{p}\right)},
\end{equation}
where ${\mathrm{o}}_{v}$ is an offset on cylindrical kernel ${\mathop{W}}_{\mathrm{cyl.}}$ for each viewpoint $v$.
The position $\mathrm{p}$ varies within in the $k\times k$ window $\mathcal{R}$.
Different from view-specific features on \figref{fig:2} (b) extracted from multi-view images, our view-specific feature benefit from structural similarity between nearby viewpoints.
Therefore, each view-specific kernel can be trained to discriminate shape variation from different viewpoints.


\subsection{Joint Category and Viewpoint Estimation}\label{sec:33}
In this section, we propose a framework to jointly estimate object category and viewpoint using the view-specific features ${F}_{v}$.
We design convolutional layers $f\left(\cdot\right)$ with parameters ${\mathop{W}}_{\mathrm{cls}}$ to produce ${N}_{v}\times \left({N}_{c}+1\right)$ score map such that ${S}_{v,c}=f\left({F}_{v};{\mathop{W}}_{\mathrm{cls}}\right)$.
Since each element of ${S}_{v,c}$ represents the probability of object belong to each category $c$ and  viewpoint $v$, the category and viewpoint can be predicted by just finding the maximum score from ${S}_{v,c}$.
However, it is not differentiable along viewpoint distribution, and only predicts discretized viewpoints.
Instead, we propose sinusoidal soft-argmax function, enabling the network to predict continuous viewpoints with periodic properties.
To obtain the probability distribution, we normalize ${S}_{v,c}$ across the viewpoint axis with a softmax operation $\sigma\left(\cdot\right)$ such that ${\mathop{P}}_{v,c}=\sigma\left({\mathop{S}}_{v,c}\right)$.
In the following, we describe how we estimate object categories and viewpoints.
\vspace{-10pt}
\paragraph{Category Classification.}
We compute the final category classification score using a weighted sum of category likelihood for each viewpoint, ${\mathop{S}}_{v,c}$, with viewpoint probability distribution, ${\mathop{P}}_{v,c}$, as follows:
\begin{equation}\label{equ:classification}
{\mathop{S}}_{c}=\sum\limits_{v=1}^{{N}_{v}}{{\mathop{S}}_{v,c}\cdot\mathop{P}}_{v,c},
\end{equation}
where ${\mathop{S}}_{c}$ represents an final classification score along category $c$.
Since the category classification is essentially viewpoint invariant, the gradient from $\mathop{S}_{c}$ will emphasize correct viewpoint's probability, while suppressing others as attention mechanism \cite{Jaderberg2015}.
It enables the back-propagation of supervisory signal along ${N}_{v}$ viewpoints.

\begin{figure}[!t]
    \centering
    {\includegraphics[width=0.99\linewidth]{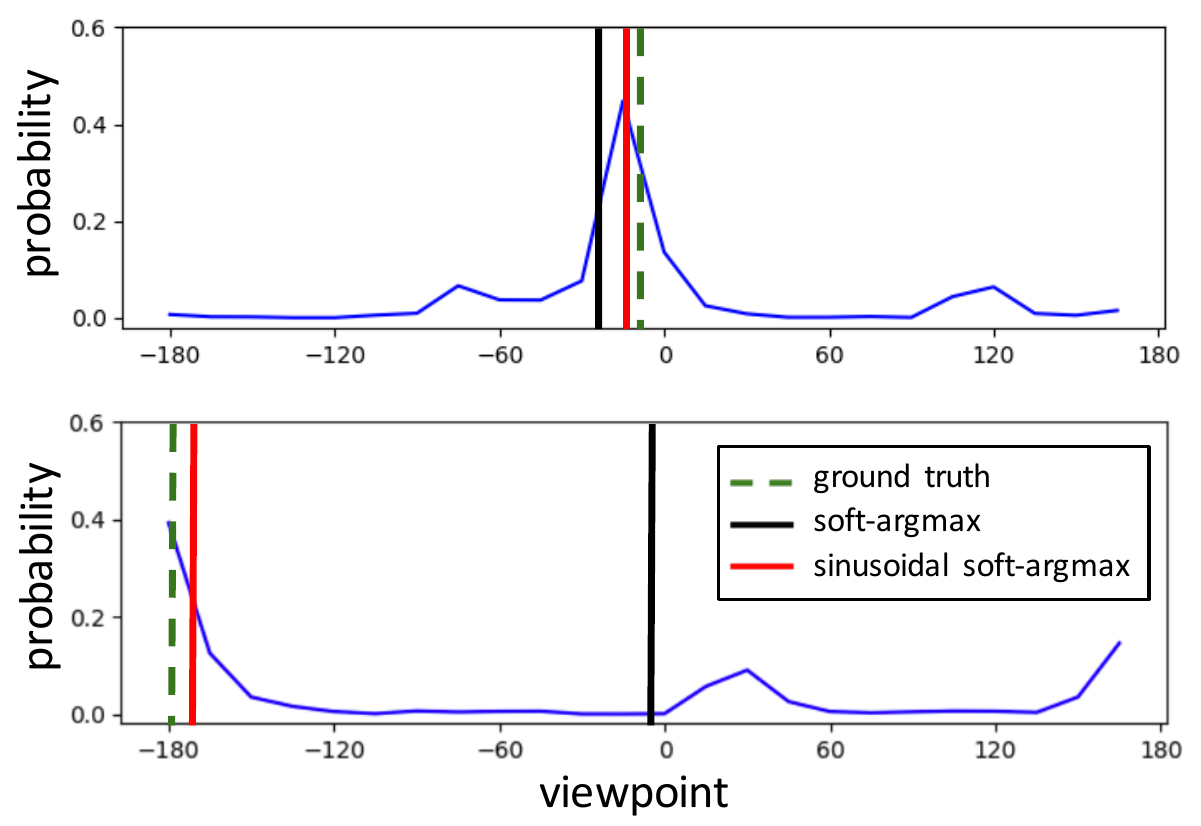}}
    \vspace{-2pt}
    \caption{Illustration of sinusoidal soft-argmax: for probability distribution of discretized viewpoint bins, sinusoidal soft-argmax enables to regress periodic viewpoint signal, while conventional soft-argmax cannot be applied.
    }
    \label{fig:4}\vspace{-14pt}
\end{figure}

\vspace{-10pt}
\paragraph{Viewpoint Estimation.}
Perhaps the most straightforward way to estimate a viewpoint within CCNs is to choose the best performing view-specific feature from predefined viewpoints to identify object category. In order to predict the continuous viewpoint with periodic properties, we further introduce a sinusoidal soft-argmax, enabling regression from ${\mathop{P}}_{v,c}$ as shown in \figref{fig:4}.

Specifically, we make use of two representative indices, $\mathrm{sin}\left({i}_{v}\right)$ and $\mathrm{cos}\left({i}_{v}\right)$, extracted by applying sinusoidal function to each viewpoint bin ${i}_{v}$ (i.e. 0\textdegree, 15\textdegree,… for ${N}_{v}=24$).
We then take sum of each representative index with its probability, followed by $\mathrm{atan2}$ function to predict object viewpoint for each class $c$ as follows:
\begin{equation}\label{equ:sinusoidalsoftargmax}
{\theta}_{c} = \mathop{\mathrm{atan2}}\left(\sum\limits_{v=1}^{{N}_{v}}{\mathop{P}}_{v,c}\sin\left({i}_{v}\right),\sum\limits_{v=1}^{{N}_{v}}{\mathop{P}}_{v,c}\cos\left({i}_{v}\right)\right),
\end{equation}
which takes advantage of classification-based approaches \cite{Tulsiani2015,Su2015} to estimate posterior probabilities, enabling better training of deep networks, while considering the periodic characteristic of viewpoints as regression-based approaches \cite{Wang2016}.
The final viewpoint estimation selects ${\theta}_{c}$ with corresponding class $c$ through category classification \equref{equ:classification}.
\vspace{-10pt}
\paragraph{Bounding Box Regression.}
To estimate fine-detailed location, we apply additional convolutional layers for bounding box regression with ${\mathop{W}}_{\mathrm{reg}}$ to produce ${N}_{v}\times{N}_{c}\times4$ bounding box offsets, denoted as ${t}_{v,c}=f\left({F}_{v};{\mathop{W}}_{\mathrm{reg}}\right)$.
Each set of 4 values encodes bounding box transformation parameters \cite{Girshick2014} from initial location for one of the ${N}_{v}\times{N}_{c}$ sets.
This leads to use different sets of boxes for each category and viewpoint bin, which can be shown as an extended version of class-specific bounding box regression \cite{Girshick2015,Ren2015}.
\vspace{-10pt}
\paragraph{Loss Functions.}
Our total loss function defined on each feature is the summation of classification loss ${L}_{\mathrm{cls}}$, bounding box regression loss ${L}_{\mathrm{reg}}$, and viewpoint estimation loss ${L}_{\mathrm{view}}$ as follows:
\begin{equation}\label{equ:multitask}
L={L}_{\mathrm{cls}}\left(c,\hat{c}\right)+[\hat{c}\geq1]\{{L}_{\mathrm{reg}}({t}_{v,c},\hat{t})
+[\hat{\theta}\neq\emptyset]{L}_{\mathrm{view}}({\theta}_{c},\hat{\theta})\},
\end{equation}
using ground-truth object category $\hat{c}$, bounding box regression target $\hat{t}$ and viewpoint $\hat{\theta}$.
Iverson bracket indicator function $\left[\cdot\right]$ evaluates to 1 when it is true and 0 otherwise.
For background, $\hat{c}=0$, there is no ground-truth bounding box and viewpoint, hence ${L}_{\mathrm{reg}}$ and ${L}_{\mathrm{view}}$ are ignored.
We train the viewpoint loss ${L}_{\mathrm{view}}$ in a semi-supervised manner, using the sets with ground-truth viewpoint ($\hat{\theta}\neq\emptyset$) for supervised learning.
For the datasets without viewpoint annotation ($\hat{\theta}=\emptyset$), ${L}_{\mathrm{view}}$ is ignored and viewpoint estimation task is trained in an unsupervised manner.
We use cross-entropy for ${L}_{\mathrm{cls}}$, and smooth L1 for both ${L}_{\mathrm{reg}}$ and ${L}_{\mathrm{view}}$, following conventional works \cite{Girshick2015,Ren2015}.

\subsection{Implementation and Training Details}\label{sec:34}

For cylindrical kernel ${\mathop{W}}_{\mathrm{cyl.}}$, we apply additional constraint to preserve a reflectinoal symmetry of 3D objects.
We first divide the parameters into four groups as front, rear, left-side, and right-side, and make the parameters of left-side and the right-side to be reflective using horizontal flip operation $h\left(\cdot\right)$ such that ${\mathop{W}}_{\mathrm{cyl.}}=[{\mathop{W}}_{\mathrm{side}},{\mathop{W}}_{\mathrm{front}},h\left({\mathop{W}}_{\mathrm{side}}\right),{\mathop{W}}_{\mathrm{rear}}]$, where parameters of each groups are concatenated horizontally.
We set the spatial resolution of ${\mathop{W}}_{\mathrm{front}}$ and ${\mathop{W}}_{\mathrm{back}}$ as $k\times1$, and ${\mathop{W}}_{\mathrm{side}}$ as $k\times\left({N}_{v}-2\right)/2$.
Therefore, ${\mathop{W}}_{\mathrm{cyl.}}$ can preserve horizontal reflectional symmetry and saves the network memory.

In order to make ${\mathop{W}}_{\mathrm{cyl.}}$ defined on a 3D space to be implemented in a 2D space, periodicity along the azimuth has to be preserved.
Therefore, we horizontally pad $k\times\lfloor{k/2}\rfloor$ of parameters from the left end to the right side using flip operation, and vice versa, where $\lfloor{\cdot}\rfloor$ denotes floor function that outputs the greatest integer less than or equal to input.
It allows ${\mathop{W}}_{\mathrm{cyl.}}$ to be used as periodic parameters.

We adopt two stage object detection framework, Faster R-CNN \cite{Ren2015} that first processes the whole image by standard fully convolutional networks \cite{He2016,Lin2017}, followed by Region Proposal Network (RPN) \cite{Ren2015} to produce a set of bounding boxes.
We then use ROI Align \cite{He2017} layer to extract fixed size feature $\mathrm{x}$ for each Region of Interest (ROI).

In both training and inference, images are resized so that the shorter side is 800 pixels, using anchors of 5 scales and 3 aspect ratios with FPN, and 3 scales and 3 aspect ratios without FPN are utilized.
2k and 1k region proposals are generated using non-maximum suppression threshold of 0.7 at both training and inference respectively.
We trained on 2 GPUs with 4 images per GPU (effective mini batch size of 8).
The backbones of all models are pretrained on ImageNet classification \cite{Deng2009}, and additional parameters are randomly initialized using \emph{He initialization} \cite{He2015}.
The learning rate is initialized to 0.02 with FPN, 0.002 without FPN, and decays by a factor of 10 at the 9th and 11th epochs.
All models are trained for 12 epochs using SGD with a weight decay of 0.0001 and momentum of 0.9, respectively.


\section{Experiments}\label{sec:4}
\begin{table}[!t]
    \centering
    \begin{tabular}{cc|cc|cc} \hline
    \multicolumn{2}{c|}{Method} & \multicolumn{2}{c|}{Cateogory} & \multicolumn{2}{c}{Viewpoint} \\ \hline
    \multicolumn{1}{c|}{${N}_{v}$} & CCNs & top-1 & top-3 & ${\mathrm{Acc}}_{\pi/6}$ & Mederr \\ \hline \hline
    \multicolumn{1}{c|}{24} &  & 0.91 & 0.97 & 0.56 & 23.5 \\
    \multicolumn{1}{c|}{18} & \checkmark & 0.95 & 0.99 & 0.63 & 17.3 \\
    \multicolumn{1}{c|}{24} & \checkmark & \bf{0.95} & \bf{0.99} & \bf{0.66} & \bf{15.5} \\
    \multicolumn{1}{c|}{30} & \checkmark & 0.94 & 0.98 & 0.63 & 17.7\\\hline
    \end{tabular}
    \vspace{6pt}
    \caption{Joint object category and viewpoint estimation performance with ground truth box on Pascal 3D+ dataset \cite{Xiang2014}.}\label{tab:1}\vspace{-10pt}
\end{table}
\subsection{Experimental Settings}\label{sec:41}
Our experiments are mainly based on maskrcnn-benchmark \cite{Massa2018} using PyTorch \cite{Paszke2017}.
We use the standard configuration of Faster R-CNN \cite{Ren2015} based on ResNet-101 \cite{He2016} as a backbone.
We implement two kinds of network, with and without using FPN \cite{Lin2017}.
For the network without using FPN, we remove the last pooling layer to preserve spatial information of each ROI feature.
We set $k=7$ following conventional works, and set ${N}_{v}=24$ unless stated otherwise.
The choice of other hyper-parameters keeps the same with the default settings in \cite{Massa2018}.

We evaluate our joint object detection and viewpoint estimation framework on the Pascal 3D+ \cite{Xiang2014} and KITTI dataset \cite{Geiger2012}.
The Pascal 3D+ dataset \cite{Xiang2014} consists of images from Pascal VOC 2012 \cite{Everingham2015} and images of subset from ImageNet \cite{Deng2009} for 12 different categories that are annotated with its viewpoint.
Note that the bottle category is omitted, since it is often symmetric across different azimuth \cite{Xiang2014}.
On the other hand, the KITTI dataset \cite{Geiger2012} consists of 7,481 training images and 7,518 test images that are annotated with its observation angle and 2D location.
For KITTI dataset, we focused our experiment on the Car object category.

\vspace{-10pt}
\paragraph{Pascal 3D+ dataset.}
In this experiment, we trained our network using the training set of Pascal 3D+ \cite{Xiang2014} (\emph{training} set of Pascal VOC 2012 \cite{Everingham2015} and ImageNet \cite{Deng2009}) for supervised learning only, denoted as CCNs, and semi-supervised learning with additional subset of \emph{trainval35k} with overlapping classes of COCO dataset \cite{Lin2014}, denoted as CCNs*.
The evaluation is done on the \emph{val} set of Pascal 3D+ \cite{Xiang2014} using Average Precision (AP) metric \cite{Everingham2015} and Average Viewpoint Precision (AVP) \cite{Xiang2014}, where we focus on AVP24 metric.
Furthermore, we also evaluate our CCNs using \emph{minival} split of COCO dataset \cite{Lin2014} using COCO-style Average Precision (AP) $@\left[0.5:0.95\right]$ and Average Recall (AR) metric \cite{Lin2014} on objects of small, medium, and large sizes.

\vspace{-10pt}
\paragraph{KITTI dataset.}
\begin{figure*}[!t]
    \centering
    \renewcommand{\thesubfigure}{}
    \subfigure[]{\includegraphics[width=0.124\linewidth]{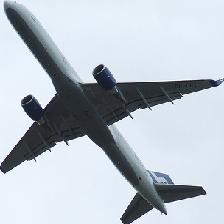}}\hfill
    \subfigure[]{\includegraphics[width=0.124\linewidth]{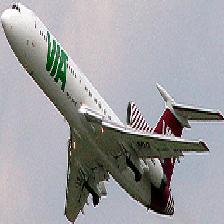}}\hfill
    \subfigure[]{\includegraphics[width=0.124\linewidth]{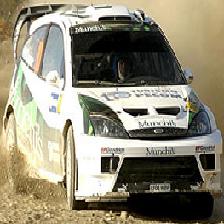}}\hfill
    \subfigure[]{\includegraphics[width=0.124\linewidth]{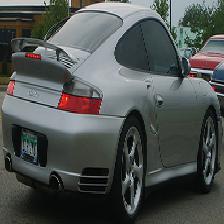}}\hfill
    \subfigure[]{\includegraphics[width=0.124\linewidth]{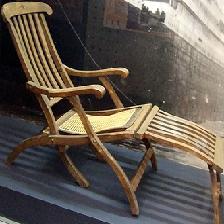}}\hfill
    \subfigure[]{\includegraphics[width=0.124\linewidth]{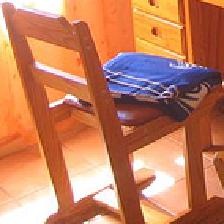}}\hfill
    \subfigure[]{\includegraphics[width=0.124\linewidth]{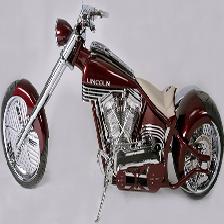}}\hfill
    \subfigure[]{\includegraphics[width=0.124\linewidth]{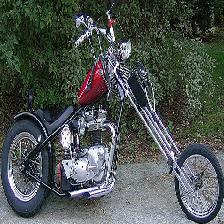}}\hfill
    \vspace{-22pt}
    \subfigure[]{\includegraphics[width=0.124\linewidth]{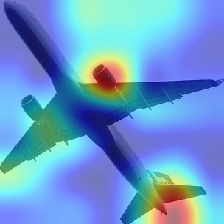}}\hfill
    \subfigure[]{\includegraphics[width=0.124\linewidth]{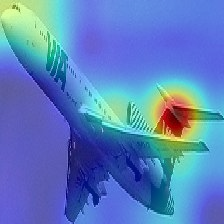}}\hfill
    \subfigure[]{\includegraphics[width=0.124\linewidth]{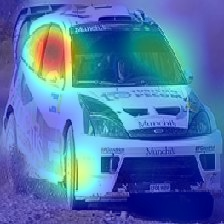}}\hfill
    \subfigure[]{\includegraphics[width=0.124\linewidth]{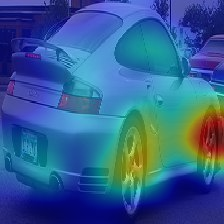}}\hfill
    \subfigure[]{\includegraphics[width=0.124\linewidth]{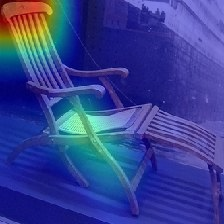}}\hfill
    \subfigure[]{\includegraphics[width=0.124\linewidth]{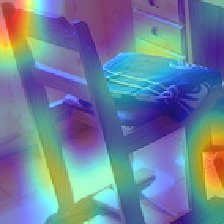}}\hfill
    \subfigure[]{\includegraphics[width=0.124\linewidth]{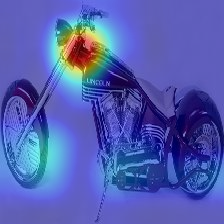}}\hfill
    \subfigure[]{\includegraphics[width=0.124\linewidth]{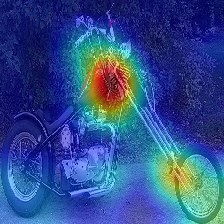}}\hfill
    \vspace{-22pt}
    \subfigure[]{\includegraphics[width=0.124\linewidth]{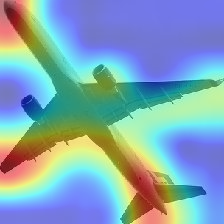}}\hfill
    \subfigure[]{\includegraphics[width=0.124\linewidth]{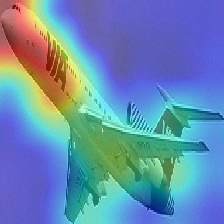}}\hfill
    \subfigure[]{\includegraphics[width=0.124\linewidth]{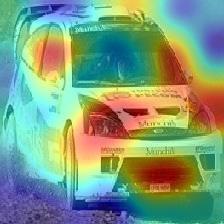}}\hfill
    \subfigure[]{\includegraphics[width=0.124\linewidth]{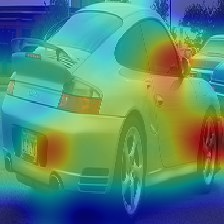}}\hfill
    \subfigure[]{\includegraphics[width=0.124\linewidth]{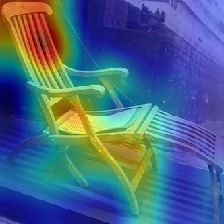}}\hfill
    \subfigure[]{\includegraphics[width=0.124\linewidth]{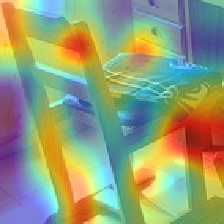}}\hfill
    \subfigure[]{\includegraphics[width=0.124\linewidth]{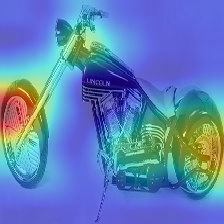}}\hfill
    \subfigure[]{\includegraphics[width=0.124\linewidth]{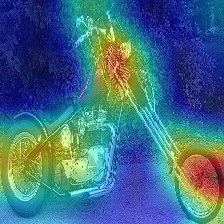}}\hfill
    \vspace{-10pt}
    \caption{Visualization of learned deep feature through Grad-CAM \cite{Selvaraju2017}: (from top to bottom) inputs, attention maps trained without CCNs, and with CCNs. Note that red color indicates attentive regions and blue color indicates suppressed regions.}
    \label{fig:5}\vspace{-10pt}
\end{figure*}

In this experiment, we followed \emph{train/val} setting of Xiang \etal \cite{Xiang2017}, which guarantees that images from the training and validation set are from different videos.
For evaluation using KITTI dataset \cite{Geiger2012}, we use Average Precision (AP) metric with $70\%$ overlap threshold (AP@IOU0.7), and Average Orientation Similarity (AOS) \cite{Geiger2012}.
Results are evaluated based on three levels of difficulty: Easy, Moderate, and Hard, which are defined according to the minimum bounding box height, occlusion, and truncation grade.


\subsection{Ablation Study}\label{sec:43}
\paragraph{Analysis of the CCNs components.}
We analyzed our CCNs with the ablation evaluations with respect to various setting of ${N}_{v}$ and the effectiveness of the proposed view-specific convolutional kernel.
In order to evaluate performance independent of factors such as mis-localization, we tackle the problem of joint category classification and viewpoint estimation with ground-truth bounding box using ResNet-101 \cite{He2016}.
For a fair comparison, a $k\times k$ view-agnostic convolutional kernels are implemented for joint object category classification and viewpoint estimation, which outputs ${N}_{c}\times{N}_{v}$ score map following conventional work \cite{Divon2018}.
In order to compare the viewpoint estimation accurately, we applied sinusoidal soft-argmax to regress the continuous viewpoint.
We evaluated the top-1 and top-3 error rates for object category classification performance, and use median error (MedErr) and ${\mathrm{Acc}}_{\pi/6}$ for viewpoint estimation performance \cite{Tulsiani2015}.

As shown in \tabref{tab:1}, CCNs have shown better performance in both object category classification and viewpoint estimation compared to the conventional method using view-agnostic kernel.
The result shows that view-specific kernel effectively leverage the complementary characteristics of the two tasks.
Since the result with ${N}_{v}=24$ has shown the best performance in both category classification and viewpoint estimation, we set ${N}_{v}=24$ for remaining experiments.
Note that the number of parameters in cylindrical kernel is $k\times\left\{({N}_{v}-2)/2+2\right\}\times{ch}_{i}=7\times13\times{ch}_{i}$, while the baseline uses $k\times k\times{ch}_{i}=7\times7\times{ch}_{i}$.
The number of additional parameters is marginal ($\sim0.01\%$) compared to the total number of network parameters, while performance is significantly improved.

\vspace{-10pt}
\paragraph{Network visualization.}
For the qualitative analysis, we applied the Grad-CAM \cite{Selvaraju2017} to visualize attention maps based on gradients from output category predictions.
We compared the visualization results of CCNs with view-specific kernel and baseline with view-agnostic kernel.
In \figref{fig:5}, the attention map of the CCNs covers the overall regions in target object, while conventional category classifier tends to focus on the discriminative part of an object.
From the observations, we conjecture that the view-specific convolutional kernel leads the network to capture the shape characteristic of object viewpoint.


\subsection{Results}\label{sec:44}
\begin{table*}[!t]
    \centering
    \begin{tabular}{c|ccccccccccc|c}
        \hline
        Method &  aero & bike & boat & bus & car & chair & dtable & mbike & sofa & train & tv & mAP \\ \hline \hline
        DPM \cite{Felzenszwalb2009} & 42.2 & 49.6 & 6.0 & 54.1 & 38.3 & 15.0 & 9.0 & 33.1 & 18.9 & 36.4 & 33.2 & 29.6 \\
        VDPM \cite{Xiang2014} & 42.2 & 44.4 & 6.0 & 53.7 & 36.3 & 12.6 & 11.1 & 35.5 & 17.0 & 32.6 & 33.6 & 29.5 \\
        DPM-VOC+VP \cite{Pepik2012} & 36.0 & 45.9 & 5.3 & 53.9 & 42.1 & 8.0 & 5.4 & 34.8 & 11.0 & 28.2 & 27.3 & 27.1 \\
        RCNN \cite{Girshick2014} & 72.4 & 68.7 & 34.0 & 73.0 & 62.3 & 33.0 & 35.2 & 70.7 & 49.6 & 70.1 & 57.2 & 56.9 \\
        Massa \etal \cite{Massa2016} & 77.1 & 70.4 & 51.0 & 77.4 & 63.0 & 24.7 & 44.6 & 76.9 & 51.9 & 76.2 & 64.6 & 61.6 \\
        Poirson \etal \cite{Poirson2016} & 76.6 & 67.7 & 42.7 & 76.1 & 59.7 & 15.5 & 51.7 & 73.6 & 50.6 & 77.7 & 60.7 & 59.3 \\
        Faster R-CNN w/ \cite{He2016} & 79.8 & 78.6 & 64.4 & 79.6 & 75.9 & 48.2 & 51.9 & 80.5 & 49.8 & 77.9 & 79.2 & 69.6 \\
        Faster R-CNN w/ \cite{Lin2017} & 82.7 & 78.3 & \bf{71.8} & 78.7 & 76.0 & \bf{50.8} & \bf{53.3} & 83.3 & 50.7 & 82.6 & 77.2 & 71.4 \\ \hline
        CCNs w/ \cite{He2016} & 82.5 & 79.2 & 64.4 & 80.3 & 76.7 & 49.4 & 50.9 & 81.4 & 48.2 & 79.5 & 78.9 & 70.2 \\
        CCNs* w/ \cite{He2016} & 82.9 & 81.4 & 63.7 & 86.6 & 79.7 & 43.6 & 51.7 & 81.6 & 52.5 & 81.0 & 82.1 & 71.5 \\
        CCNs w/ \cite{Lin2017} & 82.6 & 80.6 & 69.3 & 84.9 & 78.8 & 50.9 & 50.7 & 83.4 & 50.3 & 82.2 & 80.0 & 72.2 \\
        CCNs* w/ \cite{Lin2017} & \bf{83.7} & \bf{82.8} & 71.4 & \bf{88.1} & \bf{81.2} & 46.3 & 51.1 & \bf{85.9} & \bf{52.7} & \bf{83.8} & \bf{84.0} & \bf{73.7} \\
        \hline
    \end{tabular}
    \vspace{6pt}
    \caption{Comparison of object detection on Pascal 3D+ dataset \cite{Xiang2014}. Average Precision (AP) @IOU 0.5 is evaluated.}\label{tab:2}
\end{table*}

\begin{table*}[!t]
    \centering
    \begin{tabular}{c|ccccccccccc|c}
        \hline
        Method &  aero & bike & boat & bus & car & chair & dtable & mbike & sofa & train & tv & mAVP24 \\ \hline \hline
        VDPM \cite{Xiang2014} & 8.0 & 14.3 & 0.3 & 39.2 & 13.7 & 4.4 & 3.6 & 10.1 & 8.2 & 20.0 & 11.2 & 12.1 \\
        DPM-VOC+VP \cite{Pepik2012} & 9.7 & 16.7 & 2.2 & 42.1 & 24.6 & 4.2 & 2.1 & 10.5 & 4.1 & 20.7 & 12.9 & 13.6 \\
        Su \etal \cite{Su2015} & 21.5 & 22.0 & 4.1 & 38.6 & 25.5 & 7.4 & 11.0 & 24.4 & 15.0 & 28.0 & 19.8 & 19.8 \\
        Tulsani $\&$ Malik \cite{Tulsiani2015} & 37.0 & 33.4 & 10.0 & 54.1 & 40.0 & 17.5 & 19.9 & 34.3 & 28.9 & 43.9 & 22.7 & 31.1 \\
        Massa \etal \cite{Massa2016} & 43.2 & 39.4 & 16.8 & 61.0 & 44.2 & 13.5 & 29.4 & 37.5 & 33.5 & 46.6 & 32.5 & 36.1 \\
        Poirson \etal \cite{Poirson2016} & 33.4 & 29.4 & 9.2 & 54.7 & 35.7 & 5.5 & 23.0 & 30.3 & 27.6 & 44.1 & 34.3 & 28.8 \\
        Divon $\&$ Tal \cite{Divon2018} & \bf{46.6} & 41.1 & 23.9 & 72.6 & 53.5 & \bf{22.5} & \bf{42.6} & 42.0 & \bf{44.2} & 54.6 & 44.8 & 44.4 \\ \hline
        CCNs w/ \cite{He2016} & 39.0 & 45.9 & 22.6 & 74.5 & 54.7 & 19.6 & 38.9 & 44.2 & 41.5 & 55.3 & 46.8 & 43.9 \\
        CCNs* w/ \cite{He2016} & 39.4 & 47.0 & 23.2 & 76.6 & 55.5 & 20.3 & 39.5 & 44.5 & 41.8 & 56.1 & 45.5 & 44.5 \\
        CCNs w/ \cite{Lin2017} & 45.1 & 47.4 & 23.1 & 77.8 & 55.2 & 19.9 & 39.6 & 45.3 & 43.4 & 58.0 & 47.8 & 45.7 \\
        CCNs* w/ \cite{Lin2017} & 46.1 & \bf{48.8} & \bf{24.2} & \bf{78.0} & \bf{55.9} & 20.9 & 41.0 & \bf{45.3} & 43.7 & \bf{59.5} & \bf{49.0} & \bf{46.6} \\ \hline
    \end{tabular}
    \vspace{6pt}
    \caption{Comparison of joint object detection and viewpoint estimation on Pascal 3D+ dataset \cite{Xiang2014}.
    Average Precision with 24 discretized viewpoint bins (AVP24) is evaluated, where true positive stands with correct bounding box localization and viewpoint estimation.}\label{tab:3}\vspace{-10pt}
\end{table*}

\paragraph{Pascal 3D+ dataset.}
In the following, we evaluated our CCNs and CCNs* in comparison to the state-of-the-art methods.
Object detection methods are compared such as DPM \cite{Felzenszwalb2009}, RCNN \cite{Girshick2014}, Faster R-CNN \cite{Ren2015} with ResNet-101 \cite{He2016} and FPN \cite{Lin2017}.
Joint object detection and viewpoint estimation methods are also compared, including hand-crafted modules such as VDPM \cite{Xiang2014}, DPM-VOC+VP \cite{Pepik2012}, methods using off-the-shelf 2D object detectors for viewpoint estimation such as Su \etal \cite{Su2015}, Tulsani and Malik \cite{Tulsiani2015}, Massa \etal \cite{Massa2016}, and unified methods such as Poirson \etal \cite{Poirson2016}, Divon and Tal \cite{Divon2018}.
\begin{table}[!t]
	\centering
	\begin{tabular}{c|c|c|cccc}
		\hline
		Metric & Network & CCNs & All & S & M & L \\ \hline \hline
		\multirow{4}{*}{AP} & \multirow{2}{*}{\begin{tabular}[c]{@{}c@{}}ResNet\\\cite{He2016}\end{tabular}} &  & 34.3 & 15.5 & 28.9 & 47.3 \\
		&  & \checkmark & \bf{36.6} & \bf{17.5} & \bf{30.2} & \bf{49.6} \\ \cline{2-7} 
		& \multirow{2}{*}{\begin{tabular}[c]{@{}c@{}}FPN\\\cite{Lin2017}\end{tabular}} &  & 40.7 & 22.1 & 36.2 & 52.1 \\
		&  & \checkmark & \bf{41.8} & \bf{24.2} & \bf{38.8} & \bf{52.9} \\ \hline
		\multirow{4}{*}{AR} & \multirow{2}{*}{\begin{tabular}[c]{@{}c@{}}ResNet\\\cite{He2016}\end{tabular}} &  & 47.2 & 21.6 & 42.9 & 63.6 \\
		&  & \checkmark & \bf{49.6} & \bf{22.7} & \bf{44.1} & \bf{66.0} \\ \cline{2-7} 
		& \multirow{2}{*}{\begin{tabular}[c]{@{}c@{}}FPN\\\cite{Lin2017}\end{tabular}} &  & 54.1 & 32.6 & 51.3 & 66.5 \\
		&  & \checkmark & \bf{56.3} & \bf{33.9} & \bf{53.1} & \bf{68.3} \\ \hline
	\end{tabular}
	\vspace{2pt}
	\caption{Comparison of object detection on subset of COCO dataset \cite{Lin2014}. The COCO-style Average Precision (AP) @IOU$\in\left[0.5,0.95\right]$ and Average Recall (AR) are evaluated on objects of small (S), medium (M), and large (L) sizes.}\label{tab:4}
\end{table}
\begin{table}[!t]
    \centering
    \begin{tabular}{c|c|cccc}
    \hline
    Metric & Methods & Easy & Moderate & Hard \\ \hline \hline \cline{2-5}
    \multirow{3}{*}{AP} & Faster-RCNN \cite{Ren2015} & 82.97 & 77.83 & 66.25 \\
     & w/o CCNs & 81.74 & 76.23 & 64.19 \\
    & CCNs & \bf{86.17} & \bf{80.19} & \bf{67.14} \\ \hline
    \multirow{3}{*}{AOS} & Faster-RCNN \cite{Ren2015} & - & - & - \\
    & w/o CCNs & 79.46 & 72.92 & 59.63 \\
    & CCNs & \bf{85.01} & \bf{79.13} & \bf{63.56} \\ \hline
    \end{tabular}
    \vspace{2pt}
    \caption{Comparison of joint object detection and viewpoint estimation on \emph{val} set of KITTI dataset \cite{Geiger2012} for cars. Average Precision (AP) @IOU 0.7 is evaluated for object detection, and Average Orientation Similarity (AOS) for viewpoint estimation.}\label{tab:5}
\end{table}
\begin{figure*}[!t]
    \centering
    {\includegraphics[width=0.99\linewidth]{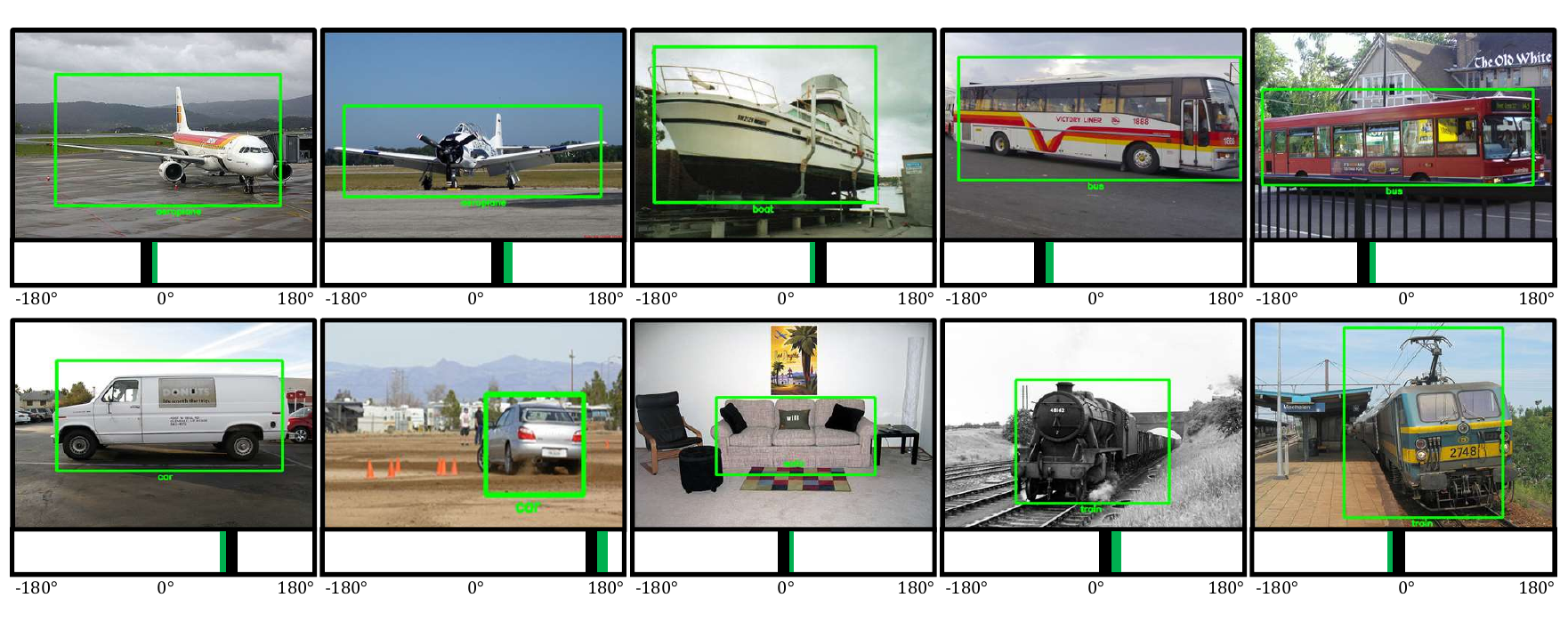}}\\
    \caption{Qualitative examples of joint object detection and viewpoint estimation on Pascal3D+ dataset \cite{Xiang2014}. The bar below each image indicates the viewpoint prediction, in green, and the ground-truth in black.}\vspace{-8pt}
    \label{fig:6}
\end{figure*}
\begin{figure*}[!t]
    \centering
    {\includegraphics[width=0.99\linewidth]{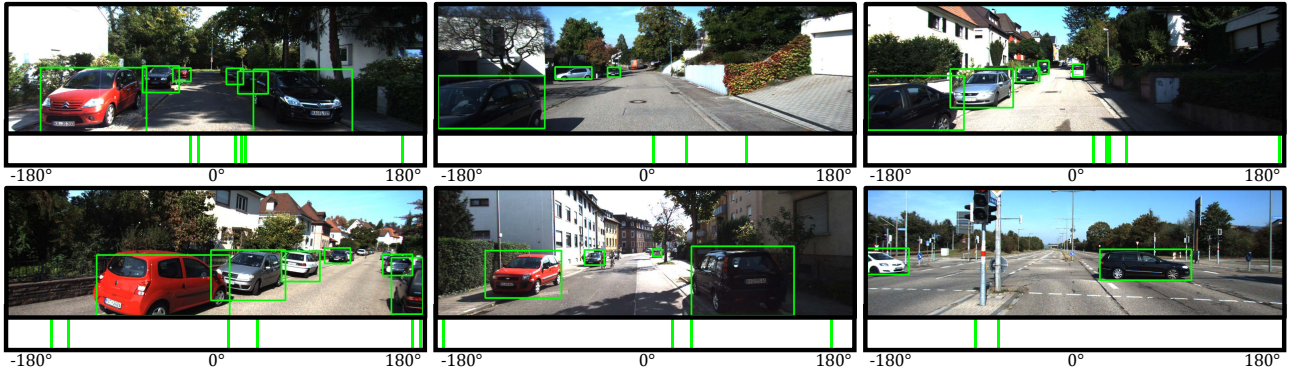}}\\
    \caption{Qualitative examples of joint object detection and viewpoint estimation on KITTI dataset \cite{Geiger2012}. The bar below each image indicates the viewpoint prediction of corresponding object in green.}
    \label{fig:7}
    \vspace{-8pt}
\end{figure*}

As shown in \tabref{tab:2} and \tabref{tab:3}, our CCNs* with FPN \cite{Lin2017} outperformed conventional methods in terms of both object detection (mAP) and joint object detection and viewpoint estimation (mAVP) on Pascal 3D+ dataset \cite{Xiang2014}.
It is noticeable that conventional methods for joint object detection and viewpoint estimation actually lowered the classification performance at \cite{Massa2016}, while ours improved the performance compared to the original Faster R-CNN \cite{Ren2015,Lin2017}.
Furthermore, our semi-supervised learning scheme using real datasets \cite{Lin2014} shows performance improvement, indicating that \equref{equ:classification} enables the supervisory signal for viewpoint estimation.
Note that other viewpoint estimation methods used synthetic images with ground-truth viewpoint annotation \cite{Su2015,Massa2016,Divon2018} or keypoint annotation \cite{Tulsiani2015}.
In \figref{fig:6}, we show the examples of our joint object detection and viewpoint estimation on Pascal 3D+ dataset \cite{Geiger2012}.

\tabref{tab:4} validates the effect of CCNs on the standard object detection dataset.
Compared to the baseline without using CCNs, object detection performance (AP) has increased by applying view-specific convolutional kernel.
Furthermore, the localization performance (AR) has also increased, indicating that our view-specific convolutional kernel can effectively encode structural information of input objects.

\vspace{-10pt}
\paragraph{KITTI dataset.}
We further evaluated our CCNs in KITTI object detection benchmark \cite{Geiger2012}.
Since the other methods aim to find 3D bounding boxes from monocular image, we conducted the experiment to validate the effectiveness of CCNs.
As shown in \tabref{tab:5}, our CCNs have shown better results compare to original Faster-RCNN \cite{Ren2015} by adapting view-specific convolutional kernel.
On the other hand, joint training of object detection and viewpoint estimation without using CCNs actually lowered the object detection performance.
This results share the same properties as previous studies \cite{Massa2016,Elhoseiny2016}, indicating that proper modeling of geometric relationship is to be determined.
In \figref{fig:7}, we show the examples of our joint object detection and viewpoint estimation on KITTI dataset \cite{Geiger2012}.
\subsection{Discussion}\label{sec:45}
Estimating the viewpoint of deformable categories is an open problem.
We thus experimented our cylindrical convolutional networks for visual recognition on rigid categories only \cite{Xiang2014}.
However, our key idea using view-specific convolutional kernel can be generalized with suitable modeling of deformable transformation (e.g., deformable convolution \cite{Dai2017}) at the kernel space.
We believe that the modeling pose or keypoint of non-rigid categories (e.g., human pose estimation) with our CCNs can be alternative to the current limitation, and leave it as future work.


\section{Conclusion}\label{sec:5}
We have introduced cylindrical convolutional networks (CCNs) for joint object detection and viewpoint estimation.
The key idea is to exploit view-specific convolutional kernels, sampled from a cylindrical convolutional kernel in a sliding window fashion, to predict an object category likelihood at each viewpoint. 
With this likelihood, we simultaneously estimate object category and viewpoint using the proposed sinusoidal soft-argmax module, resulting state-of-the-art performance on the task of joint object detection and viewpoint estimation. 
In the future, we aim to extend view-specific convolutional kernel into non-rigid categories.

{\small
\bibliographystyle{ieee_fullname}
\bibliography{egbib}
}

\end{document}